\documentclass[final,5p,times,twocolumn]{elsarticle}

\usepackage{amssymb}
\usepackage{amsmath}

\usepackage{xurl} %
\usepackage[hidelinks]{hyperref} %
\usepackage{multirow}
\usepackage{cleveref}
\usepackage{booktabs}
\DeclareMathOperator*{\argmax}{arg\,max}

\usepackage{pifont}
\newcommand{\xmark}{\ding{55}}
\usepackage{adjustbox}

\journal{Pattern Recognition Letters}

\begin{document}

\begin{frontmatter}

\title{\LARGE \bf
\includegraphics[height=1.5em]{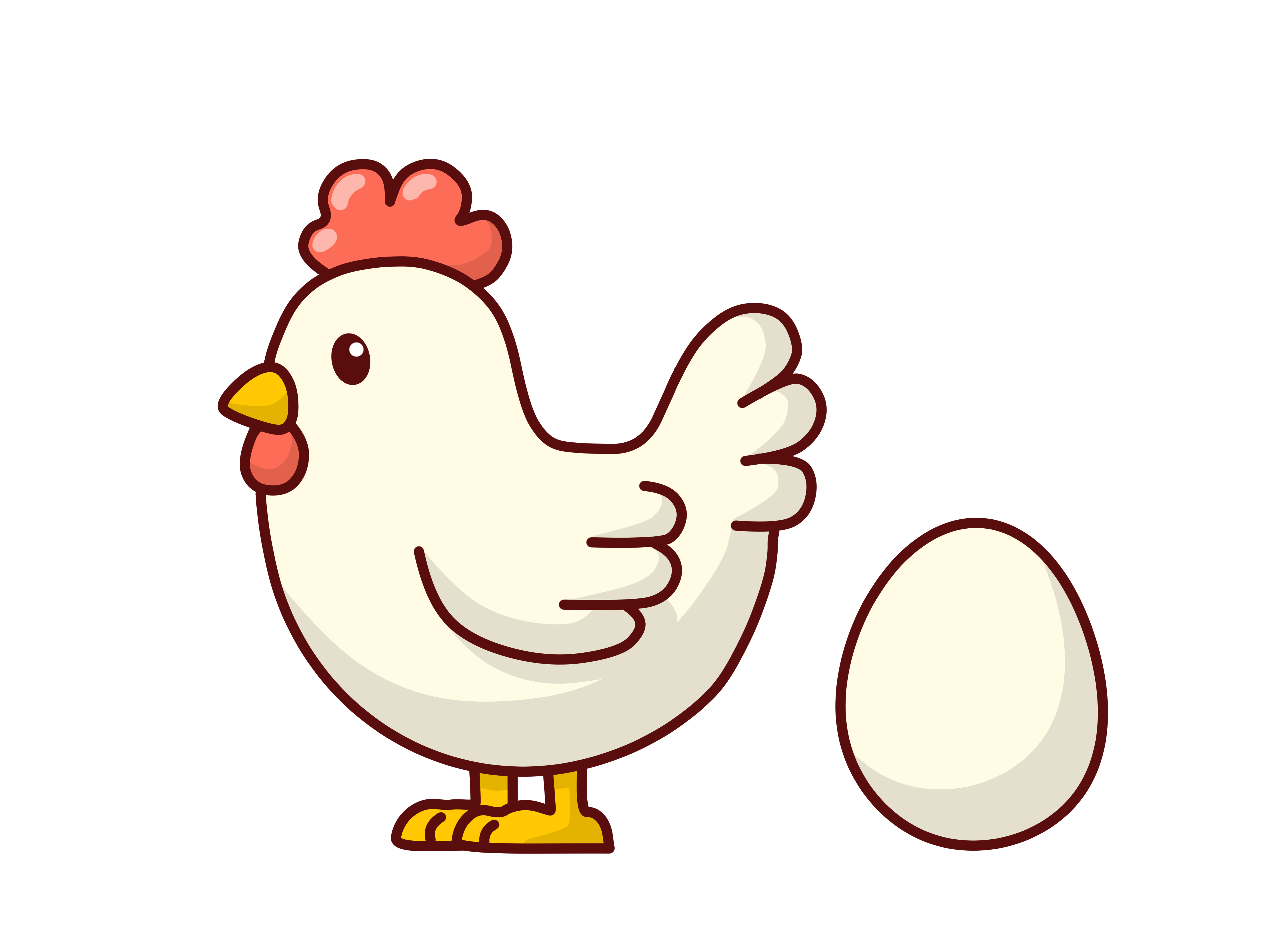}From Open-Vocabulary to Vocabulary-Free\\ Semantic Segmentation %
}

\author[label1,label2]{Klara Reichard*}
\author[label1,label3]{Giulia Rizzoli*}
\author[label1,label4]{Stefano Gasperini}
\author[label6]{Lukas Hoyer}
\author[label3]{Pietro Zanuttigh}
\author[label1]{Nassir Navab}
\author[label1,label6]{Federico~Tombari}

\affiliation[label1]{organization={Technical University of Munich}, 
                     country={Germany}}

\affiliation[label2]{organization={BMW Group},  
                     city={Munich}, 
                     country={Germany}}

\affiliation[label3]{organization={University of Padova},  
                     city={Padova}, 
                     country={Italy}}

\affiliation[label4]{organization={Munich Center for Machine Learning}, 
                     city={Munich}, 
                     country={Germany}}

\affiliation[label6]{organization={Google Zurich}, 
                     city={Zurich}, 
                     country={Switzerland}}

\begin{abstract}
Open-vocabulary semantic segmentation enables models to identify novel object categories beyond their training data. While this flexibility represents a significant advancement, current approaches still rely on manually specified class names as input, creating an inherent bottleneck in real-world applications. This work proposes a Vocabulary-Free Semantic Segmentation pipeline, eliminating the need for predefined class vocabularies. Specifically, we address the chicken-and-egg problem where users need knowledge of all potential objects within a scene to identify them, yet the purpose of segmentation is often to discover these objects. The proposed approach leverages Vision-Language Models to automatically recognize objects and generate appropriate class names, aiming to solve the challenge of class specification and naming quality. Through extensive experiments on several public datasets, we highlight the crucial role of the text encoder in model performance, particularly when the image text classes are paired with generated descriptions. Despite the challenges introduced by the sensitivity of the segmentation text encoder to false negatives within the class tagging process, which adds complexity to the task, we demonstrate that our fully automated pipeline significantly enhances vocabulary-free segmentation accuracy across diverse real-world scenarios. %
\end{abstract}

\end{frontmatter}

\section{Introduction}
\label{sec:intro}

Scene understanding is fundamental in computer vision, with segmentation playing a crucial role by providing pixel-level semantic distinctions between objects within an image. Traditionally, segmentation models were trained on closed-set, dataset-specific classes, limiting their flexibility to predefined object categories. However, as the need for a more generalized understanding of visual scenes grows, segmentation has transitioned towards open-world settings \cite{pham2018bayesian,cen2021deep}.
In open-vocabulary segmentation, models can segment any user-defined object outside the set of classes seen during training by giving as input an image and a free-form set of class names of interest \cite{xu2023side,cho2024cat}.
While open-vocabulary segmentation enables a more adaptable understanding across a range of real-world scenarios, it presents several practical challenges: %
(1) the feasibility of knowing all relevant object classes a priori -- which is also not scalable --, and (2) the quality of the class names themselves. 
Regarding (1), for open-vocabulary segmentation to work effectively, users must often specify in advance the list of the classes that the model should recognize. 
\begin{figure}[t]
    \centering
    \includegraphics[width=\columnwidth]{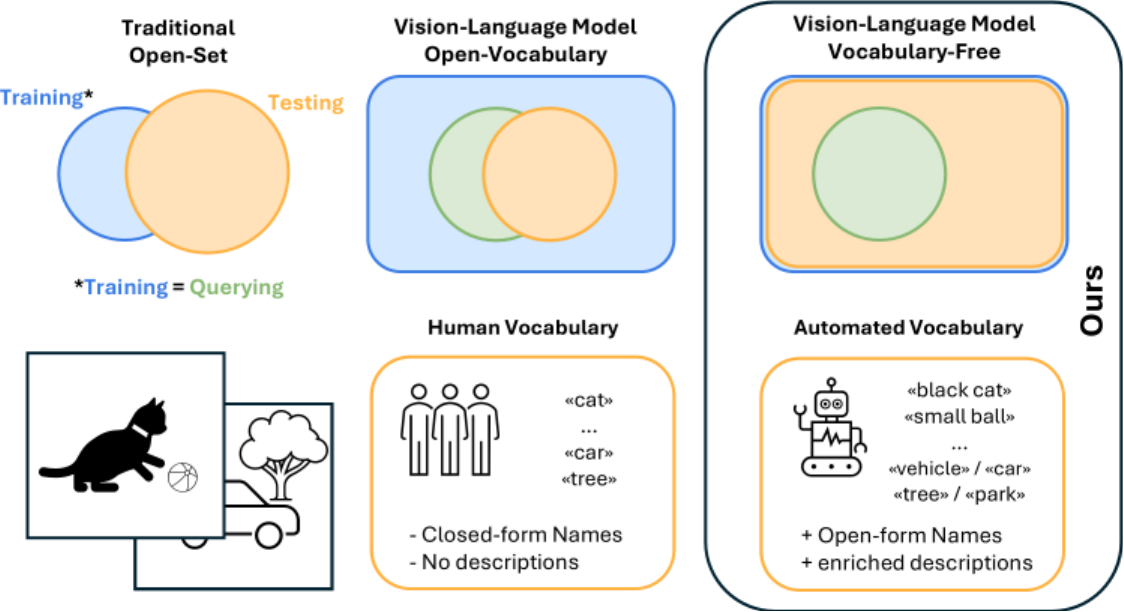}
    \caption{Traditional open-set understanding separates training and test classes, with an overlap allowing for unseen classes in testing.  In Vision-Language Models (VLMs), trained on broad, internet-scale data, open-set recognition persists but is constrained by a finite query set. In a Vocabulary-Free setting, test classes are unrestricted, allowing any concept known to the VLM, introducing more realistic, complex variations beyond predefined vocabularies.
    }
    \label{fig:teaser}
\end{figure}

This requirement can be impractical or even prohibitive in real-world applications where objects might be unexpected or difficult to enumerate. For instance, if a dataset of hotel rooms unexpectedly contains a bathtub in the middle of a room, not listing “bathtub” in the input classes could lead the model to misclassify it as something more generic or expected for the scene, like “bed” or “table.” This difficulty of predefining classes is akin to a \textbf{\textit{chicken-and-egg}} problem: users need knowledge of all potential objects within a scene to identify them, yet the purpose of segmentation is often to discover these very objects. This limitation restricts open-vocabulary segmentation’s applicability in settings where out-of-context or out-of-distribution objects appear.
Moreover, for (2) the quality of class names themselves often adds to these challenges. As highlighted in recent works, existing annotations frequently use imprecise or overly general terms, which may not align with human categorizations \cite{huang2024renovating}. Large datasets typically involve multiple annotators, who may assign different names for the same visual concept, leading to inconsistent or non-intuitive labels. For example, a class like ``vehicle'' may sometimes be labeled as ``car'' or ``transport,'' depending on annotators' preferences. This inconsistency can hinder segmentation models’ ability to learn precise object distinctions, as they may interpret general labels more broadly than intended, or miss specific instances altogether.

To address these limitations, we propose a fully automated pipeline that first performs image tagging to recognize objects, and then generates descriptive context when needed, before proceeding with segmentation. More in detail, we integrate the original annotations of the training and validation datasets with language model capabilities to create a more comprehensive and adaptive system. Differently from prompt-based methods - like SAM \cite{kirillov2023segment} - or other user-guided frameworks such as traditional open-vocabulary approaches \cite{cho2024cat,shin2024towards}, our pipeline operates in a completely automated manner, leveraging both human-verified labels (in training only) and the model's broader semantic understanding to handle unexpected objects and scenarios.
An overview of the task is shown in \Cref{fig:teaser}.
Our contribution can be decomposed into different stages:
\begin{itemize}
    \item First, we propose a two-stage pipeline for Vocabulary-Free segmentation, composed of an image tagging module and a class-specific decoder based on VLMs. Due to its simplicity and performance superiority over existing approaches, it serves as a benchmark for the community to accelerate future research on VSS.
    \item Next, under the assumption of a perfect image tagger, we explore the effects of providing enriched textual inputs to the text encoder in the segmentation model.
    \item  Finally, we assess the impact of unreliable tagging on segmentation performances, particularly the effect of undetected objects - false negatives - and wrong ones - false positives.
\end{itemize}

\section{Related Work}
\label{sec:related}
\textbf{Open-Set Understanding:}
The open-set task, first introduced by Scheirer et al. \cite{scheirer2012toward}, challenges the conventional closed-set paradigm commonly assumed in image recognition. In closed-set models, the testing phase only includes samples from a predefined set of classes known to the model during training. Conversely, open-set recognition addresses scenarios where samples can belong to previously unknown classes that were not present during training. This requires models to both recognize and reject instances from unfamiliar classes, ensuring robustness against unknown inputs. The open-set framework has seen extensive study across multiple areas in computer vision, such as image classification~\cite{bendale2016towards, vaze2022open, yoshihashi2019classification, oza2019c2ae, perera2020generative, chen2021adversarial,zhang2020hybrid}, %
object detection~\cite{han2022expanding, miller2020uncertainty, zhou2023open}, %
and image segmentation~\cite{hwang2021exemplar, pham2018bayesian, cen2021deep}.%

\textbf{Open-vocabulary Semantic Segmentation (OVSS):}
Recent advances in vision-language models (VLMs) such as CLIP~\cite{radford2021learning} have demonstrated that robust, transferable visual representations can be effectively learned from large-scale datasets using only weakly structured natural language descriptions.
Initially adopted in image-level tasks like classification, VLMs leverage both visual and textual embeddings to recognize a diverse set of classes. %
By aligning image features with semantic concepts in a shared space, VLMs achieve a form of zero-shot learning that allows them to identify new classes at test time, offering a flexible framework for generalization \cite{liu2024open,xie2024sed,cho2024cat}. 
Although open-vocabulary learning presents an appealing solution for handling arbitrary classes, scaling this approach to accommodate an ever-growing set of classes poses significant challenges. In theory, a VLM could achieve perfect generalization if its query set contained every conceivable class label. However, as demonstrated by Miller et al. \cite{miller2025open}, adding more classes to the query set does not lead to better performance. In fact, increasing the number of class labels introduces a greater likelihood of misclassifications, leading to degraded model accuracy. This degradation occurs because, as more classes are added, the semantic space becomes increasingly crowded, causing overlaps that make accurate distinctions between classes harder to achieve. 
To tackle scalability, one solution is training class-free models \cite{shin2024towards}, while distinguishability can be improved by enhancing the textual descriptions of the classes \cite{ma2024open,jiao2024collaborative}. However, all these approaches assume that the inference label set is predefined and available at inference time.%

\textbf{Vocabulary-Free Semantic Segmentation (VSS):} Recent research in VSS has focused on developing end-to-end solutions while reducing bias from  ground truth data annotations. The majority of current methods decompose the task into a class-agnostic mask generation and a class association (Mask2Tag). Zero-Seg \cite{rewatbowornwong2023zero} and TAG \cite{kawano2024tag} leverage DINO \cite{caron2021emerging} to generate the masks, followed by CLIP-based \cite{radford2021learning} embedding generation for class association. Zero-Seg processes these embeddings %
following ZeroCap \cite{tewel2021zero} to obtain textual classes, while TAG matches them against an external database. 
Conversely, CaSED \cite{conti2024vocabulary} identifies potential classes by querying an external caption database using a pre-trained VLM. %
Similarly, Auto-Seg \cite{ulger2024autovocabularysemanticsegmentation} %
fine-tunes a captioning model to extract class names at multiple scales, followed by a second stage where an open-vocabulary model generates segmentation masks, with predictions remapped to ground truth classes using LLMs. While these approaches demonstrate promising results, they do not fully explore how this pipeline decomposition impacts model performance, nor do they investigate methods to enrich the textual information in the CLIP encoder. We address these limitations by providing a comprehensive analysis of the text encoder's role and exploring techniques to enhance visual-language understanding through richer textual representations. Moreover, we rigorously test the sensitivity of CLIP to tagger errors, evaluating how inaccuracies in image tagging propagate and impact the final segmentation performance.

\begin{figure*}[t]
    \centering
    \includegraphics[width=.9\textwidth]{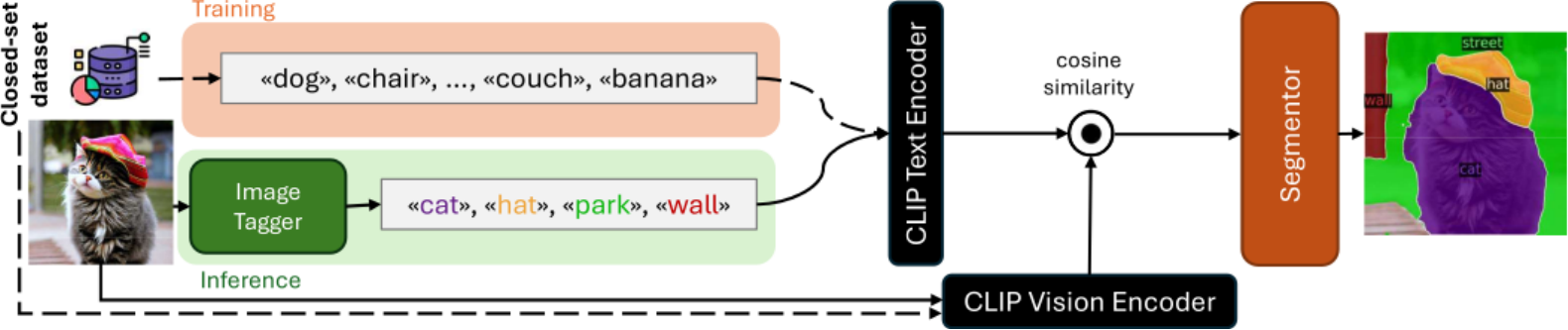}
    \caption{Mask2Tag systems rely on class-agnostic masking models, where masks are associated with names. Oppositely, in our Tag2Mask pipeline, the image tagger supervises the text encoder, while the similarity between the text and vision embeddings produces the segmentation prediction. The segmentation module benefits from pretraining on a highly curated, close-set dataset.}
    \label{fig:method}
\end{figure*}

\section{Best practice for VSS}
\label{ch:method}
In this section, we describe our experimental framework for analyzing the role of each component in Vocabulary-Free Semantic Segmentation. %
Unlike previous methods that rely on zero-training segmentation modules with limited class transfer capabilities \cite{conti2024vocabulary} or implement simplistic mask2tag pipelines \cite{rewatbowornwong2023zero,kawano2024tag}, our approach leverages the complementary strengths of two state-of-the-art models: a text-based image tagger for semantic supervision and an open-set segmentation model, as shown in \Cref{fig:method}.
Specifically, we introduce a system that utilizes a text tagger to provide class-level semantic information, employs a segmentation model with global conceptual knowledge from CLIP, and fine-tunes the model on a training dataset to learn precise segmentation skills. We first formalize the Vocabulary-Free Semantic Segmentation (VSS) problem (\Cref{sec:formulation}), then detail how we integrate these models in our training setup (\Cref{sec:training}). Finally, we describe the evaluation protocol for assessing segmentation quality (\Cref{sec:assignment}).
The approach emphasizes the novelty of the system's strategy rather than claiming to propose entirely new components, highlighting the integration of existing state-of-the-art models and explaining the specific improvements over previous methods through its innovative class-aware strategy. By combining these approaches, we provide a comprehensive framework that advances segmentation capabilities in open-set scenarios.

\subsection{Problem Formulation}\label{sec:formulation}
In the standard open-vocabulary segmentation (OVSS) setting, given an image \( I \), the set of class names (labels) \( \mathcal{C} = \{c_1, c_2, \dots, c_N\} \) is known a priori. In contrast, our approach assumes that the class name set \( \mathcal{C} \) is pre-defined during training time, yet at inference time, we associate the image with an open-ended set of words predicted by a vision-language model, rather than the fixed set \( \mathcal{C} \).

Specifically, the goal of Vocabulary-Free Semantic Segmentation (VSS) is to segment an image \( x \) into regions corresponding to semantic concepts without any prior knowledge of the semantic categories \( C \). Instead of using a fixed vocabulary of classes, we operate directly within a vast semantic space \( S \), which includes all possible semantic concepts.
In the VSS setup, the function \( f \) maps the image space \( X \) to a semantic map in \( S^{(H \times W)}\), and each pixel is assigned a label from the large semantic space \( S \). The function is defined as \( f: X \to S^{(H \times W)} \).
At test time, the function \( f \) relies solely on the input image \( x \) and a broad repository of semantic concepts approximating \( S \). This enables zero-shot segmentation, where the model can segment an image into novel categories  not part of the training data.

To implement this, our pipeline first performs image tagging to recognize objects present in the image. This step generates a set of descriptive tags that may include both known concepts and unexpected objects. Next, we leverage the broader semantic understanding of the vision-language model to provide additional contextual information about these detected objects, going beyond the limited set of predefined class names. Finally, we use this %
textual representation to guide the segmentation process, allowing our model to handle a dynamic and open-ended set of visual concepts without requiring retraining for new categories. The pipeline of the approach is shown in \Cref{fig:method}.

This task is inherently challenging due to the vast size of the semantic space \( S \), which is much larger than typical predefined label sets. For example, while the largest segmentation benchmark for open vocabulary has roughly 900 classes, resources like the vast web or a large-scale knowledge base contain millions of potential semantic concepts, covering a much broader range of categories. The model must differentiate between concepts across diverse domains, and handle long-tailed distributions and ambiguous regions in images. The VSS approach also needs to address fine-grained recognition and segmentation of objects, which is difficult without predefined class labels.

\subsection{Two-stage Pipeline}\label{sec:training}
While theoretically, a training dataset may introduce biases due to its limited label lexicon, training-free or class-free open-vocabulary methods often yield inferior segmentation results compared to trained approaches \cite{shin2024towards}. Additionally, vocabulary-free methods present a more challenging scenario, as classes are not predefined and may be overlooked if the tagger fails to recognize them. Nevertheless, we opted for a trained segmentation backbone, by %
adopting the open-vocabulary training scheme. 
During inference, tagging is guided by a pre-trained VLM to predict potential classes within the images. %
The segmentation baseline %
follows.

\textbf{Segmentation framework:} 
Initially, dense image embeddings \( D^V \in \mathbb{R}^{(H \times W) \times d} \) are generated using the CLIP image encoder \( \Phi^V \), where \( H \) and \( W \) represent the height and width of the image, respectively, and \( d \) is the embedding dimension. Concurrently, each class name in  \( \mathcal{C} \) is used in a template \( T \) - "A photo of  a \{class name\}" - to produce the text embeddings \( D^L \in \mathbb{R}^{N \times d} \) that are generated using a CLIP text encoder \( \Phi^L \), where \( N \) is the number of classes.
A cost volume \( C \in \mathbb{R}^{(H \times W) \times N} \) is computed by calculating the cosine similarity between each spatial location in the image embeddings and each class embedding. The cost volume is defined as:
\begin{equation}
    C_{i,n} = \frac{D^V(i) \cdot D^L(n)}{|D^V(i)| \, |D^L(n)|}
\end{equation}
where \( i \) indexes the spatial locations and \( n \) indexes the classes.
The algorithm  \cite{cho2024cat} then performs \( K \) iterations of feature aggregation to refine the initial cost embedding \( F \). Each iteration consists of two steps: 

1. \textbf{Spatial aggregation:}
\begin{equation}
    F'(:,n) = T^{\text{sa}}([F(:,n); P^V(D^V)])
\end{equation}

A spatial aggregation transformer \( T^{\text{sa}} \) is applied to each class channel, incorporating guidance from the projected image embeddings \( P^V(D^V) \). 

2. \textbf{Class aggregation:}
\begin{equation}
    F''(i,:) = T^{\text{ca}}([F'(i,:); P^L(D^L)])
\end{equation}
A class aggregation transformer \( T^{\text{ca}} \) is applied to each spatial location, incorporating guidance from the projected text embeddings \( P^L(D^L) \).
After each iteration, the refined features \( F'' \) are used as input for the next iteration.
At the end of the \( K \) iterations of feature aggregation, an upsampling decoder is applied to the final refined features \( F \). This decoder produces the segmentation prediction for all \( N \) classes.

\subsection{Evaluation Assignment}
\label{sec:assignment}
The predictions of the image tagger are not necessarily aligned with the dataset labels as the latter does not allow for synonyms or coarse-to-fine assignments (e.g., if "tower" is not a label, it should be interpreted as "building"). While not strictly necessary for functionality, a remapping process is required to enable a fairer evaluation consistent with those used in prior studies \cite{kawano2024tag,rewatbowornwong2023zero}. This results in two types of mean Intersection over Union (mIoU) metrics: a hard assignment, calculated when the predictions from the image tagger are exactly the same as the ground truth class names, and a soft mIoU, calculated using the remapping process.
For the soft mIoU, the goal is to map each predicted class name $p_i$ to the most appropriate ground truth class name $c_j$. If a predicted class name $p_i$ is already present in the set of ground truth class names $\mathcal{C}$, it is directly assigned to the corresponding $c_j$.
For the remaining predicted class names, we compute sentence embeddings using a pre-trained model, e.g., Sentence-BERT \cite{reimers2019sentence}. Then, for each $p_i \in \mathcal{P}$ and $c_j \in \mathcal{C}$, we calculate the cosine similarity between their embeddings. The ground truth class name $c^*$ that maximizes this similarity is assigned to the predicted class name $p_i$:
\begin{equation}
    c^* = \argmax_{c_j \in \mathcal{C}} \text{sim}(\Phi^L(p_i), \Phi^L(c_j))
\end{equation}
where $\text{sim()}$ is the cosine similarity between the embeddings.
This remapping approach allows the model's predictions to be better aligned with the ground-truth labels, even when the predicted and ground-truth texts do not exactly match. Note that, if the highest cosine similarity score is below a threshold $T_\text{SBERT}$, the match is discarded.

\begin{table}[t]
\centering
\caption{Results over the benchmark datasets. The mIoU is reported. %
}
\label{tab:sota_results}
\resizebox{\columnwidth}{!}{
\begin{tabular}{ccccccc}
\toprule
\textbf{Method} & \begin{tabular}[c]{@{}c@{}}Inference\\ Vocab. \end{tabular} & A-847 & PC-459 & A-150 & PC-59 & VOC-20 \\ \midrule
SAN \cite{xu2023side} & \checkmark & 12.4 & 15.7 & 27.5 & 53.8 & 94.0 \\
AttrSeg \cite{ma2024open} & \checkmark & -- & -- & -- & 56.3 & 91.6 \\
SCAN \cite{liu2024open} & \checkmark & 14.0 & 16.7 & 30.8 & \textbf{58.4} & \textbf{97.0} \\
EBSeg \cite{shan2024open} & \checkmark & 13.7 & 21.0 & 30.0 & 56.7 & 94.6 \\
SED \cite{xie2024sed} & \checkmark & 11.4 & 18.6 & 31.6 & 57.3 & 94.4 \\
CAT-Seg \cite{cho2024cat} & \checkmark & \textbf{16.0} & \textbf{23.8} & \textbf{31.8} & 57.5 & 94.6 \\ \midrule
CaSED + SAM \cite{conti2024vocabulary} & \xmark & -- & -- & 6.1 & 7.5 & 13.7 \\
CaSED + SAN \cite{conti2024vocabulary} & \xmark & -- & -- & 7.2 & 15.5 & 26.9 \\
DenseCaSED \cite{conti2024vocabulary} & \xmark & -- & -- & 8.6 & 13.4 & 20.5 \\
\textbf{Chick.-and-egg} (CaSED) & \xmark & 3.2 & 4.4 & 9.7 & \textbf{23.1} & \textbf{47.6} \\
\textbf{Chick.-and-egg} (RAM) & \xmark & \textbf{3.7} & \textbf{7.1} & \textbf{15.6} & 23.0 & 47.5  \\
\bottomrule
\end{tabular}
}
\end{table}

\section{Experiments}
\label{ch:results}

We conduct a comprehensive experimental analysis to investigate how different components affect VSS performance.
First, we evaluate the proposed two-stage approach on standard benchmarks to establish the baseline (\Cref{sec:benchmark}). We then present an in-depth analysis of the text encoder's behaviour and its impact on segmentation quality (\Cref{sec:text}). To better understand the relationship between the two-stages, we examine the image tagging accuracy and its influence on the segmentation task (\Cref{sec:tagging}). Finally, we study how different assignment thresholds in the evaluation protocol affect the reported performance (\Cref{sec:thresholds}).

\textbf{Implementation Details:}
The model is trained on the COCO-Stuff dataset \cite{caesar2018coco}, which contains 118k annotated images across 171 categories, following \cite{cho2024cat}. All results are based on CLIP \cite{radford2021learning} with a ViT-B/16 backbone. The image encoder and cost aggregation module are trained with per-pixel binary cross-entropy loss. 
The training parameters follow \cite{cho2024cat}. The batch size is 4, and models are trained for 80k iterations.
We performed image tagging and instance description using a frozen VLM model not trained on the testing dataset. More in detail, we examined the robustness of two models RAM \cite{zhang2024recognize} and Llava-1.6  \cite{liu2024llavanext}.

\textbf{Test Datasets:} The evaluation covers several datasets to ensure comprehensive testing. We used ADE20K \cite{zhou2019semantic} with both 150 and 847 class configurations, Pascal Context \cite{mottaghi2014role} with 59 and 459 class setups, and Pascal VOC \cite{everingham2010pascal} with its 20 classes. 

\begin{figure*}[t]
    \centering
    \resizebox{\textwidth}{!}{%
    \begin{tabular}{@{}ccccc@{}}

        \includegraphics[width=0.25\textwidth]{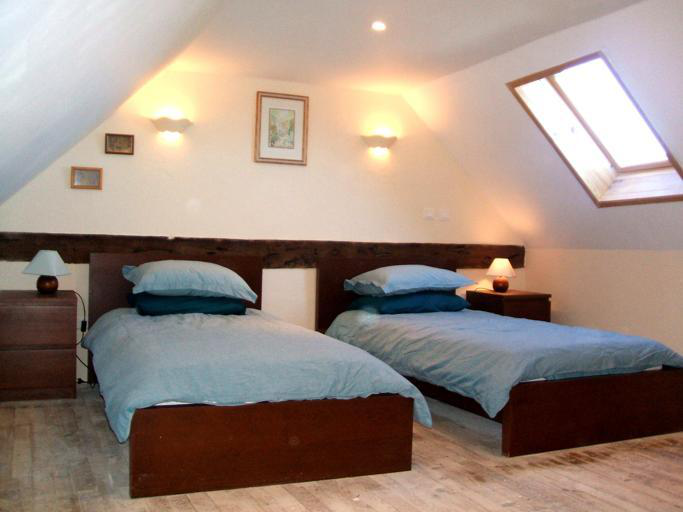} &
        \includegraphics[width=0.25\textwidth]{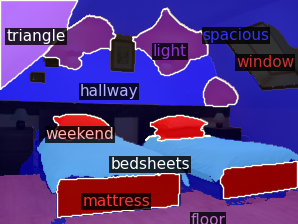} &
        \includegraphics[width=0.25\textwidth]{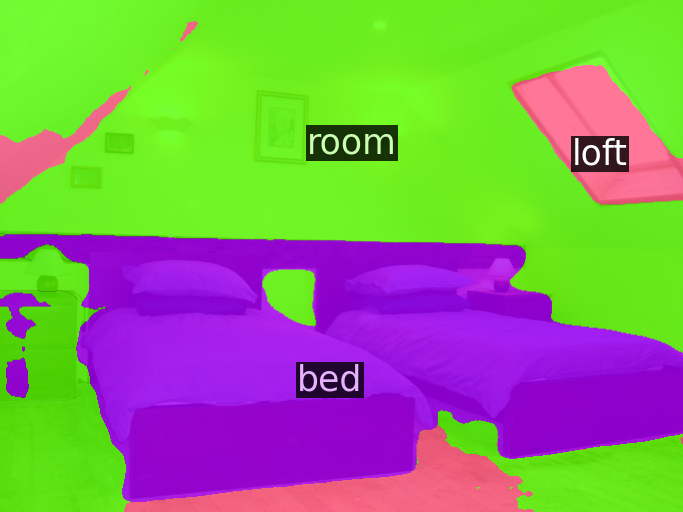} &
        \includegraphics[width=0.25\textwidth]{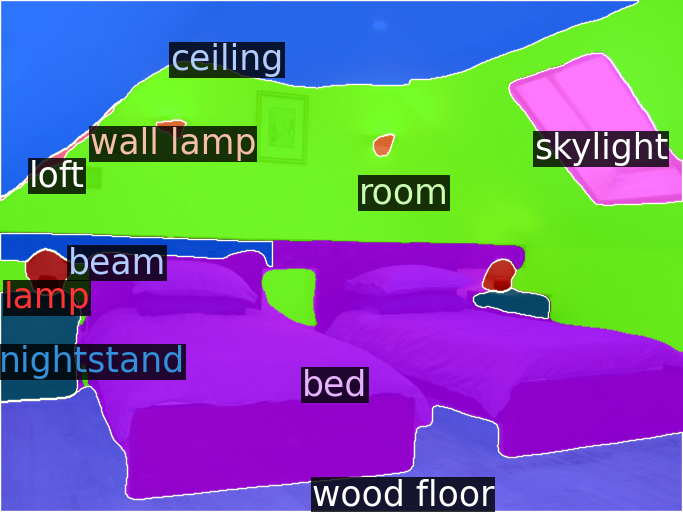} &
        \includegraphics[width=0.25\textwidth]{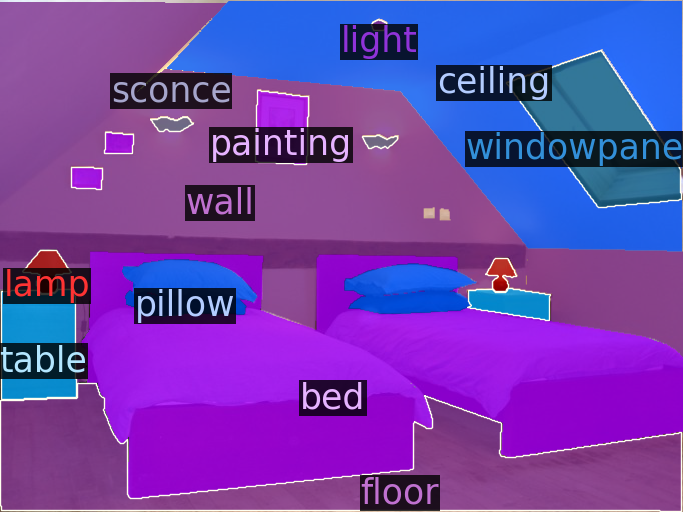} \\[0.2cm]

        \includegraphics[width=0.25\textwidth] {fig/qualitative_new/ADE_val_00000683_img.png} &
        \includegraphics[width=0.25\textwidth] {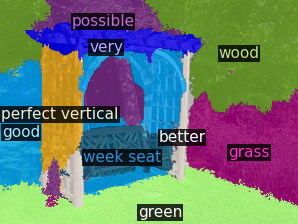} &
        \includegraphics[width=0.25\textwidth]{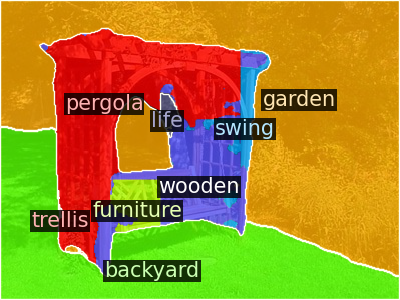} &
        \includegraphics[width=0.25\textwidth]{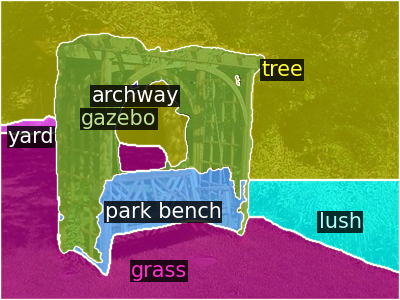} &
        \includegraphics[width=0.25\textwidth]{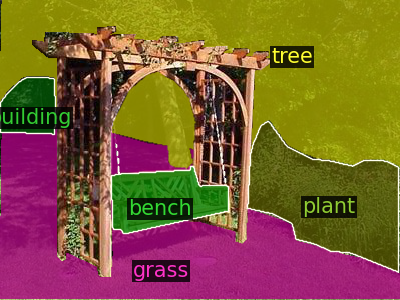} \\[0.2cm]

        \includegraphics[width=0.25\textwidth] {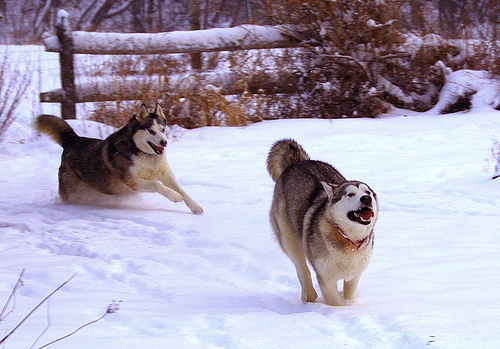} &
        \includegraphics[width=0.25\textwidth] {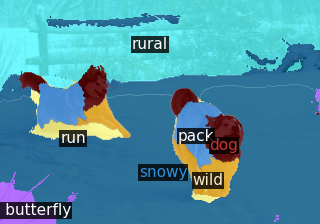} &
        \includegraphics[width=0.25\textwidth]{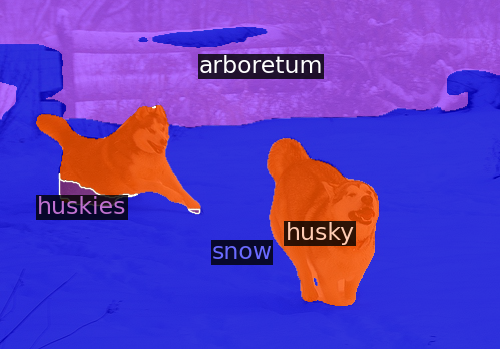} &
        \includegraphics[width=0.25\textwidth]{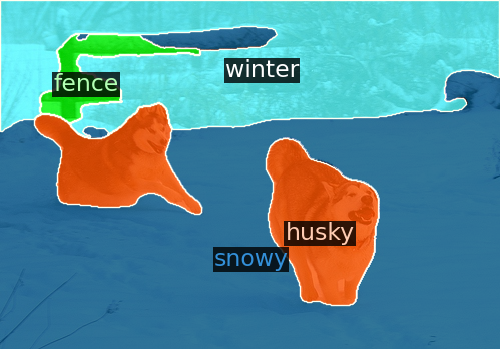} &
        \includegraphics[width=0.25\textwidth]{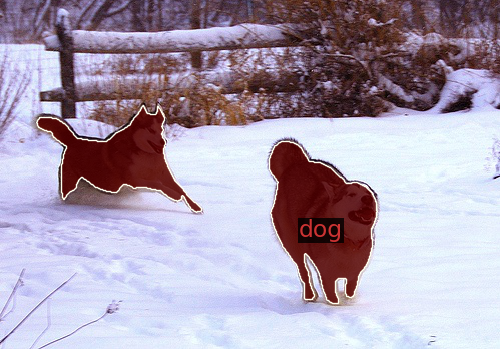} \\[0.2cm]

        \textbf{Image} &
        \textbf{ZeroSeg} & 
        \textbf{Chicken-and-egg} (CaSED) & 
        \textbf{Chicken-and-egg} (RAM) & 
        \textbf{GT}
    \end{tabular}
    }
    \caption{Comparison of segmentation results across ZeroSeg \cite{rewatbowornwong2023zero} and \textbf{Chicken-and-Egg} (CaSED \cite{conti2024vocabulary} and RAM \cite{zhang2024recognize}), and ground-truth labels.}
    \label{fig:sota_qualitative_comparison}
\end{figure*}

\begin{table*}[t]
\centering
\caption{Results over the benchmark datasets by using soft assignment. † Results come from their original work. * mapped with Llama-2 \cite{ulger2024autovocabularysemanticsegmentation} rather than Sentence-BERT \cite{reimers2019sentence}. The soft assignment has threshold zero (i.e., all the words are assigned to a class in the evaluation vocabulary).}
\label{tab:mapping_results}
\resizebox{.99\textwidth}{!}{%
    \begin{tabular}{l|cc|cc|cccccccccc}
    \toprule
    \multirow{2}{*}{\textbf{Method}} & \multirow{2}{*}{\begin{tabular}[c]{@{}c@{}}\textbf{Vision}\\ \textbf{Backbone}\end{tabular}} & \multirow{2}{*}{\textbf{Stages}} &\multicolumn{2}{c}{\textbf{Components}} & A-847 & PC-459 & A-150 & PC-59 & VOC-20 \\
     & & & Tagging & Segmentation & &&&&& \\ \midrule
    Zero-Seg† \cite{rewatbowornwong2023zero} & ViT-B/16 & Mask2Tag & CLIP+GPT-2 & DINO & -- & -- & -- & 11.2 & 8.1 \\
    Auto-Seg† \cite{ulger2024autovocabularysemanticsegmentation} & ViT-L/16 & Tag2Mask & BLIP-2 & X-Decoder & 5.9* & -- & -- & 11.7* & \textbf{87.1}* \\
    TAG† \cite{kawano2024tag} & ViT-L/14 & Mask2Tag &CLIP+DB & DINO & -- & -- & 6.6 & 20.2 & 56.9 \\
    \textbf{Chicken-and-egg} (CaSED) & ViT-B/16 & Tag2Mask & CLIP+DB & CAT-Seg & 4.3 & 3.1 & 7.8 & \textbf{27.9} & 82.3 \\
    \textbf{Chicken-and-egg} (RAM) & ViT-B/16 & Tag2Mask & CLIP+Swin & CAT-Seg & \textbf{6.7} & \textbf{8.0} & \textbf{18.8} & 27.8 & 81.8 \\
        \bottomrule
    \end{tabular}
}
\end{table*}

\subsection{Benchmark Evaluation}\label{sec:benchmark}
We first conducted a comprehensive benchmark evaluation comparing existing approaches to establish a strong foundation for VSS and identify the most promising direction. This analysis served two key purposes: (1) to understand the current state-of-the-art performance in VSS and (2) to determine which baseline architecture would be the most suitable.

\textbf{Quantitatives:} \Cref{tab:sota_results} compares the mIoU across the Open-Vocabulary benchmarks. The proposed pipeline outperforms the previous VSS methods by a constant margin in all the datasets. To better accommodate VSS methods, they adopt a class remapping strategy that reduces penalization in cases where an exact class match is not found. This approach is reflected in \Cref{tab:mapping_results}, where the soft evaluation assignment takes place as described in \Cref{sec:assignment}.

\textbf{Qualitatives:} As shown in \Cref{fig:sota_qualitative_comparison}, the current approach fills the gap between the predictions and original dataset labels without a predefined vocabulary, offering finer-grained details across diverse scenarios (indoor and outdoor). %
The maps obtained suggest that current evaluation metrics might be overly pessimistic about the qualitative performance of the results. This issue arises from dataset limitations, where many instances struggle to find appropriate matches (e.g., in the third image, "husky" instead of "dog"). Mask2Tag methods like ZeroSeg \cite{rewatbowornwong2023zero} tend to over-segment the instances, getting improper text matches. On the other hand, Chicken-and-egg with CaSED tends to limit the number of predicted tags, %
while coupled with RAM it reaches the best compromise.

\subsection{Segmentatation Analysis} \label{sec:text}
\textbf{Perfect Tagger:} Our empirical results on the OVSS task - presented in \Cref{tab:gt_labels} - revealed that providing only image-specific text labels, %
rather than the entire vocabulary, during training led to improved segmentation performance when applying the same adjustment at inference. Although having access to inference labels is unrealistic, this setup represents the best achievable performance if tagger predictions were 100\% accurate. 
More in detail, in \Cref{tab:gt_labels}, the set of class names is defined for each batch as \(\mathcal{C}_b \subset \mathcal{C}\) during training, where \(\mathcal{C}_b\) represents the batch-specific subset of the entire class vocabulary \(\mathcal{C}\), dynamically selected based on the batch's unique context or requirements. This subset approach allows the model to focus on relevant classes without being overwhelmed by the entire vocabulary. However, we observed no gain when the text labels in inference are predicted from an image tagger. Nevertheless, this represents the upper bound currently obtainable with the state-of-the-art open-vocabulary method \cite{cho2024cat}. Moreover, we show in \Cref{tab:attvsadj} that when performing inference on perfect predictions (100\% accuracy from the tagger) we can boost performance by providing additional textual information.
\begin{table}[t]
\centering
\caption{\textbf{Comparison using CAT-Seg \cite{cho2024cat}, using ground truth classes as text embeddings at different stages}, where $T$ represents training and $I$ represents inference. The mIoU is reported on ADE-20K (A)\cite{zhou2019semantic}, Pascal Context (PC)\cite{mottaghi2014role}, and Pascal VOC (VOC) \cite{everingham2010pascal}.
}
\label{tab:gt_labels}
\resizebox{\columnwidth}{!}{%
\begin{tabular}{ccccccccc}
\toprule
\textbf{Method} & \multicolumn{2}{c}{\textbf{\begin{tabular}[c]{@{}c@{}}Only GT\\ Text Labels\end{tabular}}} & COCO & A-847 & PC-459 & A-150 & PC-59 & VOC-20 \\ \cmidrule{2-3}
 & T & I &  &  &  &  &  &  \\ \midrule
Base &  &  & 47.11 & 11.95 & 18.95 & 31.78 & 57.20 & 95.30 \\
L.Bound & \checkmark &  & 43.73 & 10.89 & 16.63 & 30.29 & 55.99 & 94.20 \\
U.Bound (I) &  & \checkmark & 56.15 & 12.38 & 18.38 & 45.53 & 69.77 & \textbf{95.87} \\
U.Bound & \checkmark & \checkmark & \textbf{64.03} & \textbf{13.98} & \textbf{24.04} & \textbf{51.21} & \textbf{72.79} & 94.38 \\
\bottomrule
\end{tabular}
}
\end{table}

\begin{table}[t]
\centering
\caption{All methods are based on \cite{cho2024cat}, changing textual descriptors, while performing inference on GT classes. (a)-(c) are trained using the predicted VLM information on COCO dataset.
}
\label{tab:attvsadj}
\resizebox{\columnwidth}{!}{%
\begin{tabular}{ccccccccc}
\toprule
\textbf{Method} & \multicolumn{2}{c}{\textbf{VLM input}} & COCO & A-847 & PC-459 & A-150 & PC-59 \\ \cmidrule{2-3}
& $Image$& $Text$ & & & \\ \midrule
Baseline \cite{cho2024cat} & & & 56.15 & 12.38 & 18.38 %
& 45.53 & 69.77 &\\ %
(a) Caption & \checkmark & & 58.17 & 12.71 & 17.07 %
& 47.09 & 71.04 &\\ %
(b) Class Adjectives & & \checkmark & 62.33 & 14.96 & 19.13 & 48.77 & 60.47 \\ %
(c) Instance Adjectives & \checkmark & \checkmark & \textbf{65.13} & \textbf{15.40} & \textbf{23.20} & \textbf{54.43} & \textbf{72.04} \\ %
\bottomrule
\end{tabular}
}
\end{table}

\begin{table*}[t]
\centering
\caption{Prompts for different algorithms for \cref{tab:attvsadj} results.}
\label{tab:prompt}
\resizebox{.9\textwidth}{!}{%
\begin{tabular}{cc}
\toprule
\textbf{\begin{tabular}[c]{@{}c@{}}Description\\ Level\end{tabular}} & \textbf{Prompts} \\ \midrule
Class & \begin{tabular}[c]{@{}c@{}}1. "Please group the classes in this list $<$dataset-class-list$>$ into groups of classes that are similar to each other \\ meaning they could be confused in an image. Every class should be in one group and only in one group. \\ Make sure there are no classes from the original list missing in your grouping. \\ This is an example of how the output should look: dog, cat, kitten, bird -- couch, desk, sofa, lamp -- knife, fork, plate"\\ 2. " The classes in the group are: $<$group$>$. Please generate a short list of adjectives for each class \\ that describe how the object looks in an image. The adjectives should be distinctive within each group meaning that \\ the same attribute should not appear for two classes in the same group. Generate at least one adjective for each class. \\ This is an example how the output should look. {giraffe: [tall, brown, spotted, yellow], tree: [tall, green], armchair: [comfortable]}\end{tabular}\\ \midrule
Instance & \begin{tabular}[c]{@{}c@{}}"The objects in the image are: $<$dataset-class-list$>$. Please generate a short list of adjectives\\ for each object that describes how the object looks in the image. \\ This is an example of how the output should look. \{giraffe: {[}tall, brown, spotted, interacting{]}, tree: {[}tall, green, leafy{]}\}"\end{tabular} \\ \bottomrule
\end{tabular}
}
\end{table*}

\textbf{Aiding Text Encoder with Descriptions:} Previous works \cite{ma2024open} used adjectives with the assumption to find the common class features that better describe each class. For example, a "dalmata" could be described as "a white dog with black spots". However, in typical recognition tasks, the categories are much broader, such as simply "dog", and a "dog" could be described very differently in terms of color and size. Hence, AttrSeg \cite{ma2024open} have focused on training strategies to find the optimal set of descriptions that could enhance class distinguishability while still being able to represent each class. While this approach has merit, it can result in the loss of fine-grained details. For instance, a "table" or "hat" could be of any size or color, and even a "wall" that is typically "white" could be "bricked" or some other texture.
Zhao et al. \cite{zhao2024gradient} experimented on CLIP's ability to identify different types of object attributes, including shape, material, color, size, and position. 
For shape and material attributes, CLIP showed a certain but limited knowledge, with the heat maps highlighting partial correct attention on obvious objects, but also exhibiting false positive and false negative errors. For color attributes, the results further verified that CLIP has a good ability to distinguish different colors.
For comparative attributes like size and position, CLIP produced some erroneous results, demonstrating that it relies more on the primary object (e.g., "cube", "red") rather than the comparative attribute (e.g., "small", "left"). Overall, their analysis suggests that CLIP has advantages with common perceptual attributes.
Therefore, we adopted a pre-trained VLM to find the corresponding descriptions given each image and its specific set of class names - the text labels of each image-, and we tried to enforce general language descriptions.
The prompts used are shown in \Cref{tab:prompt}. For generating captions, we employed the BLIP-2 model \cite{li2023blip} without any query input, whereas for the multimodal model, we utilized Llava-1.6 \cite{liu2024llavanext}. These models were selected because they both incorporate CLIP as their text encoder.
The text embedding of the captions is employed as a query within an additional cross-attention module, linking it to the embeddings of the classes. In the case of the adjectives, they are sampled and used within the template "A photo of a \{adjective\} \{class name\}".
We report the results in table \Cref{tab:attvsadj}, where adding image-specific content results beneficial, specially for large numbers of classes.
It is important to notice that, when using predicted labels from the image tagger or applying the complete set of image labels during inference, we did not observe the same benefit. %
In the VSS scenario, ambiguities with other classes are largely resolved during the CLIP segmentation stage by directly predicting the image's content using the image tagger. However, misclassifications may still occur at this stage, a behaviour explored in the next paragraph.
\begin{table}[t]
\centering
\caption{Class recognition accuracy of different VLMs with $T_\text{SBERT}$=$0.0$. \\ * using vocabulary.
\# FN = average number of missed classes, \# FP = average number of classes predicted but not in the ground truth.}
\label{tab:accuracy}
\resizebox{.5\textwidth}{!}{%
\begin{tabular}{ccccccccccc}
\toprule
\multirow{2}{*}{\textbf{\begin{tabular}[c]{@{}c@{}}Predicted\\ Classes\end{tabular}}} & \multirow{2}{*}{\textbf{\begin{tabular}[c]{@{}c@{}}Mapping\\ Model\end{tabular}}} & \multicolumn{3}{c}{A-150} & \multicolumn{3}{c}{PC-59} & \multicolumn{3}{c}{VOC-20} \\ \cmidrule{3-11} 
 &  & Acc & \#FP & \#FN & Acc & \#FP & \#FN & Acc & \#FP & \#FN \\ \midrule
CaSED & - & 10 & 10.7 & 7.8 & 22 & 9.3 & 4.0 & 50 & 9.5 & 0.9 \\
CaSED & SBERT & 23 & 7.4 & 6.8 & 42 & 5.4 & 3.1 & 84 & 4.2 & 0.3 \\
Llava-1.6 & - & 26 & 4.9 & 6.3 & 29 & 3.7 & 3.5 & 53 & \textbf{3.5} & 0.8 \\
Llava-1.6 & SBERT & 39 & \textbf{2.6} & 5.2 & 47 & \textbf{1.8} & 2.7 & 91 & 1.9 & 0.2 \\
RAM & - & 34 & 10.4 & 5.9 & 41 & 11.8 & 3.1 & 68 & 12.2 & 0.5 \\
RAM & SBERT & \textbf{46} & 5.7 & \textbf{4.8} & \textbf{61} & 5.4 & \textbf{2.2} & \textbf{96} & 4.8 & \textbf{0.1} \\
\midrule
RAM* & - & 79 & 16.7 & 1.95 & 80 & 6.2 & 1.1 & 97 & 1.5 & 0.1  \\
\bottomrule
\end{tabular}
}
\end{table}

\subsection{Image Tagging Analysis}\label{sec:tagging}
\begin{figure}
    \centering
    \includegraphics[width=.9\columnwidth, trim=0cm 0.55cm 0cm 0.75cm, clip]{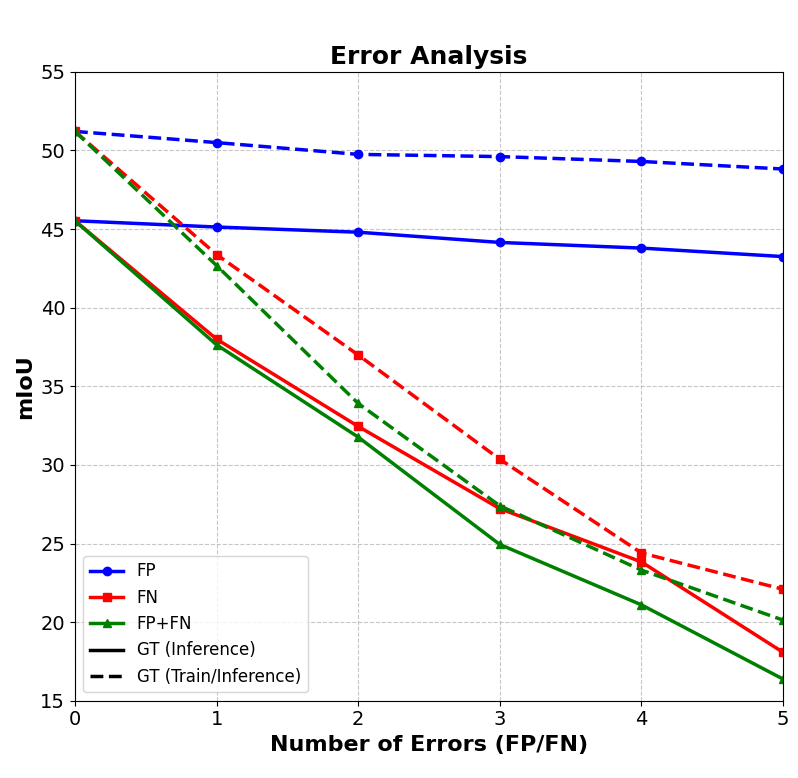}
    \caption{Simulating missing classes or adding wrong ones over the OVSS baseline by assuming the labels are known at inference time.}
    \label{fig:miss_vs_false_sim}
\end{figure}

In \Cref{tab:accuracy}, we investigated various image tagging methods to understand how different types of errors affect the sensitivity of the segmentation module, particularly the text encoder since we use the tags as input to CLIP. We evaluated the impact of three architectures: a training-free method, CaSED \cite{conti2024vocabulary}, a multi-step trained method, RAM \cite{zhang2024recognize}, and a general-purpose multimodal model, Llava \cite{liu2024llavanext}.
CaSED %
uses a pre-trained vision-language model and an external database to extract candidate categories and assign the image to the best match. 
On the other hand, RAM %
generates large-scale image tags through automatic semantic parsing, followed by training a model to annotate images using both captioning and tagging tasks. A data engine then refines these annotations, and the model is retrained on this enhanced data, with final fine-tuning on a higher-quality subset.
\Cref{tab:accuracy} shows that using SBERT for evaluation avoids discarding words merely due to the absence of an exact match with the chosen word by the annotators. RAM achieves the best overall results across the evaluated datasets. In the table, the performance of Llava \cite{liu2024llavanext} %
demonstrates the versatility of powerful vision-language architectures. Note that the current baseline, RAM, does not reach a perfect accuracy even when the whole list of desired classes (i.e., non-vocabulary free), hence this represents the current limitation of such an approach. 
Furthermore, compared to CaSED, RAM demonstrates higher class recognition accuracy, but with more false positives on average. To investigate this further, we examined in \Cref{fig:miss_vs_false_sim} how the model is influenced by simulating a drop rate and false positives on top of the ground truth text classes in each image. In the table, the false positives are randomly selected from the vocabulary set. The influence of false negatives deeply influences the performance, while introducing false positives only leads to marginal degradation. These results confirm why RAM outperforms current alternatives: it has the fewest misclassifications, despite having a higher rate of false positives.

\subsection{Evaluation Assignment Thresholds} \label{sec:thresholds}
\begin{figure}
    \centering
    \includegraphics[width=.9\columnwidth, trim=0cm 0.55cm 0cm 0.75cm, clip]{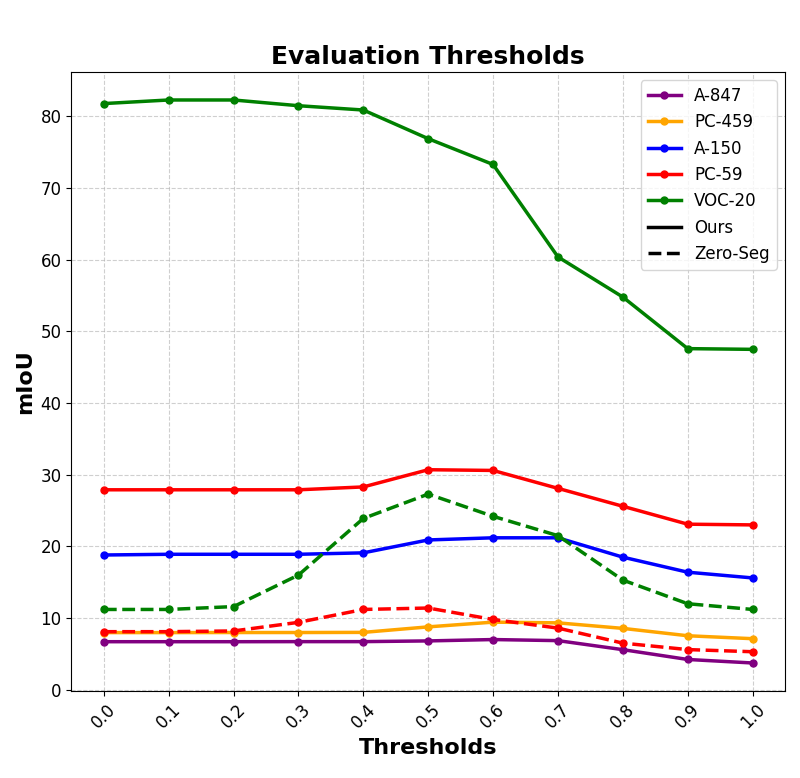}
    \caption{Ablation over different thresholds for the evaluation mapping.%
    }
    \label{fig:thresh}
\end{figure}

In \Cref{fig:thresh} we show the effect of providing different values for $T_\text{SBERT}$. Unlike Zero-Seg \cite{rewatbowornwong2023zero}, we did not observe a consistent trend in the optimal threshold across datasets. Respectively, $0.6-0.7$ for A-847, PC-459 and A-150, $0.5$ for PC-59 and $0.1$ for VOC-20. 
Our findings suggest that as the number of classes increases, we need to be more confident in the assignment, hence a higher threshold leads to a better score.

\section{Conclusions}
The paper proposes a Vocabulary-Free Semantic Segmentation (VSS) pipeline that addresses the limitations of current open-vocabulary segmentation approaches. Existing methods require users to specify object classes in advance, which is impractical for real-world scenarios with unexpected objects. VSS approaches leverage Vision-Language Models to automatically recognize objects and generate appropriate class names, eliminating the need for predefined vocabularies. Our experiments show the proposed pipeline significantly improves segmentation accuracy over the previous methods, particularly for out-of-context objects. Moreover, we demonstrated that the performance of the model is heavily influenced by the text encoder, which plays a crucial role in its ability to generalize. Its effectiveness is closely linked to the alignment between the actual image content and the generated class descriptions, with integrated class information showing promise for enhancing recognition. However, our results also reveal that the sensitivity of the text encoder to overlooked objects represents a challenge for achieving a flawless segmentation system.

\section*{Acknowledgment}
This manuscript is currently under consideration at Pattern Recognition Letters.
The Bayerische Motoren Werke (BMW) and the Deutscher Akademischer Austauschdienst (DAAD) partially supported this work.

\bibliographystyle{elsarticle-num}
\bibliography{references}

\end{document}